\documentclass[letterpaper]{article} 
\usepackage{aaai20}  
\usepackage{times}  
\usepackage{helvet} 
\usepackage{courier}  
\usepackage[hyphens]{url}  
\usepackage{graphicx} 
\usepackage{graphicx}  
\usepackage{algorithmic}
\usepackage{algorithm}
\usepackage{amsmath}
\usepackage{amsthm}
\usepackage{amssymb,bbm}
\newtheorem{theorem}{Theorem}
\newtheorem*{theorem*}{Theorem}

\usepackage{tikz}
\usepackage{subfigure}

\title{Options of Interest: Temporal Abstraction with Interest Functions}
\author{Khimya Khetarpal,\textsuperscript{\rm 1,\rm 2} Martin Klissarov,\textsuperscript{\rm 1,\rm 2} Maxime Chevalier-Boisvert,\textsuperscript{\rm 2,\rm 3} \\
\Large \textbf{Pierre-Luc Bacon,\textsuperscript{\rm 4} Doina Precup\textsuperscript{\rm 1,\rm 2,\rm 5}} \\   
\textsuperscript{\rm 1}McGill University, \textsuperscript{\rm 2}Mila, $^3$Universite de Montreal,  \\ \textsuperscript{\rm 4}Stanford University, \textsuperscript{\rm 5}Google DeepMind \\
khimya.khetarpal@mail.mcgill.ca, martin.klissarov@mail.mcgill.ca \\
maxime.chevalier-boisvert@mila.quebec, plbacon@cs.stanford.edu, dprecup@cs.mcgill.ca}

\begin{document}
\maketitle
\begin{abstract}
Temporal abstraction refers to the ability of an agent to use behaviours of controllers which act for a limited, variable amount of time. The options framework describes such behaviours as consisting of a subset of states in which they can initiate, an internal policy and a stochastic termination condition. However, much of the subsequent work on option discovery has ignored the initiation set, because of difficulty in learning it from data. We provide a generalization of initiation sets suitable for general function approximation, by defining an interest function associated with an option. We derive a gradient-based learning algorithm for interest functions, leading to a new interest-option-critic architecture. We investigate how interest functions can be leveraged to learn interpretable and reusable temporal abstractions. We demonstrate the efficacy of the proposed approach through quantitative and qualitative results, in both discrete and continuous environments.
\end{abstract}

\section{Introduction}
Humans have a remarkable ability to acquire skills, and knowing when to apply each skill plays an important role in their ability to quickly solve new tasks. In this work, we tackle the problem of learning such skills in reinforcement learning (RL). AI agents which aim to achieve goals are faced with two difficulties in large problems: the depth of the lookahead needed to obtain a good solution, and the breadth generated by having many choices. The first problem is often solved by providing shortcuts that skip over multiple time steps, for example, by using macro-actions~\cite{hauskrecht1998hierarchical}. The second problem is handled by restricting the agent's attention at each step to a reasonable number of possible choices. Temporal abstraction methods aim to solve the first problem, and a lot of recent literature has been devoted to this topic~\shortcite{konidaris2011autonomous,mann2015approximate,kulkarni2016hierarchical,mankowitz2016adaptive,bacon2017option,machado2017laplacian}. We focus specifically on the second problem: learning how to reduce the number of choices considered by an RL agent.

In classical planning, the early work on STRIPS~\cite{Fikes1972} used preconditions that had to be satisfied before applying a certain action. Similar ideas can also be found in later work on macro-operators~\cite{Korf1983} or the Schema model~\cite{Drescher1991}. In RL, the framework of options~\cite{sutton1999between} uses a similar concept, \textit{initiation sets}, which limit the availability of options (i.e. temporally extended actions) in order to deal with the possibly high cost of choosing among many options. Moreover, initiation sets can also lead to options that are more \textit{localized}~\cite{konidaris2007building}, which can be beneficial in transfer learning. For example, in continual learning~\shortcite{ring1997child}, specialization is key to both scaling up learning in large environments, as well as to ``protecting" knowledge that has already been learned from forgetting due to new updates.

The option-critic architecture~\cite{bacon2017option} is a gradient-based approach for learning options in order to optimize the usual long-term return obtained by an RL agent from the environment. However, the notion of initiation sets originally introduced in~\citeauthor{sutton1999between} \shortcite{sutton1999between} was omitted from~\citeauthor{bacon2017option} \shortcite{bacon2017option} due to the difficulty of learning sets with gradient-based methods. We propose a generalization of initiation sets to \textit{interest functions}~\cite{Sutton2016,White17}. We build from the fact that a set can be represented through its membership function. Interest functions are a generalization of membership functions which allows smooth parameterization. Without this extension, determining suitable initiation sets would necessitate a non-differentiable, search-based approach. 

\textbf{Key Contributions:} We generalize initiation sets for options to \textit{interest functions}, which are differentiable, and hence easier to learn. We derive a gradient-based learning algorithm capable of learning all components of options end-to-end. The resulting interest-option-critic architecture generates options that are specialized to different regions of state space. We demonstrate through experiments in both discrete and continuous problems that our approach generates options that are useful in a single task, interpretable and reusable in multi-task learning.\begin{figure*}[h!]
   \centering
    \subfigure[]{\label{fig:earlystage}\includegraphics[width=0.25\textwidth]{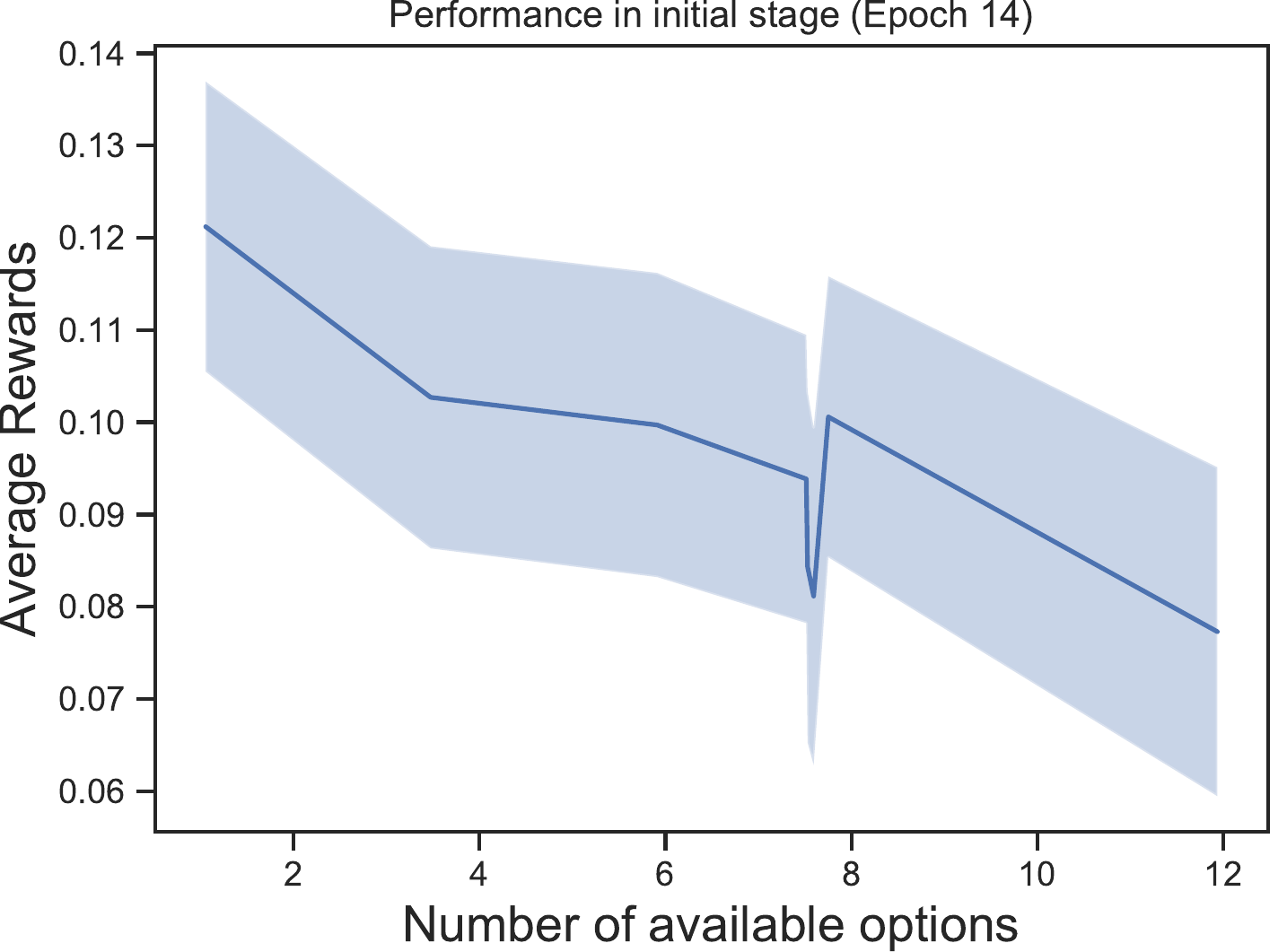}}
    \subfigure[]{\label{fig:latestage}\includegraphics[width=0.25\textwidth]{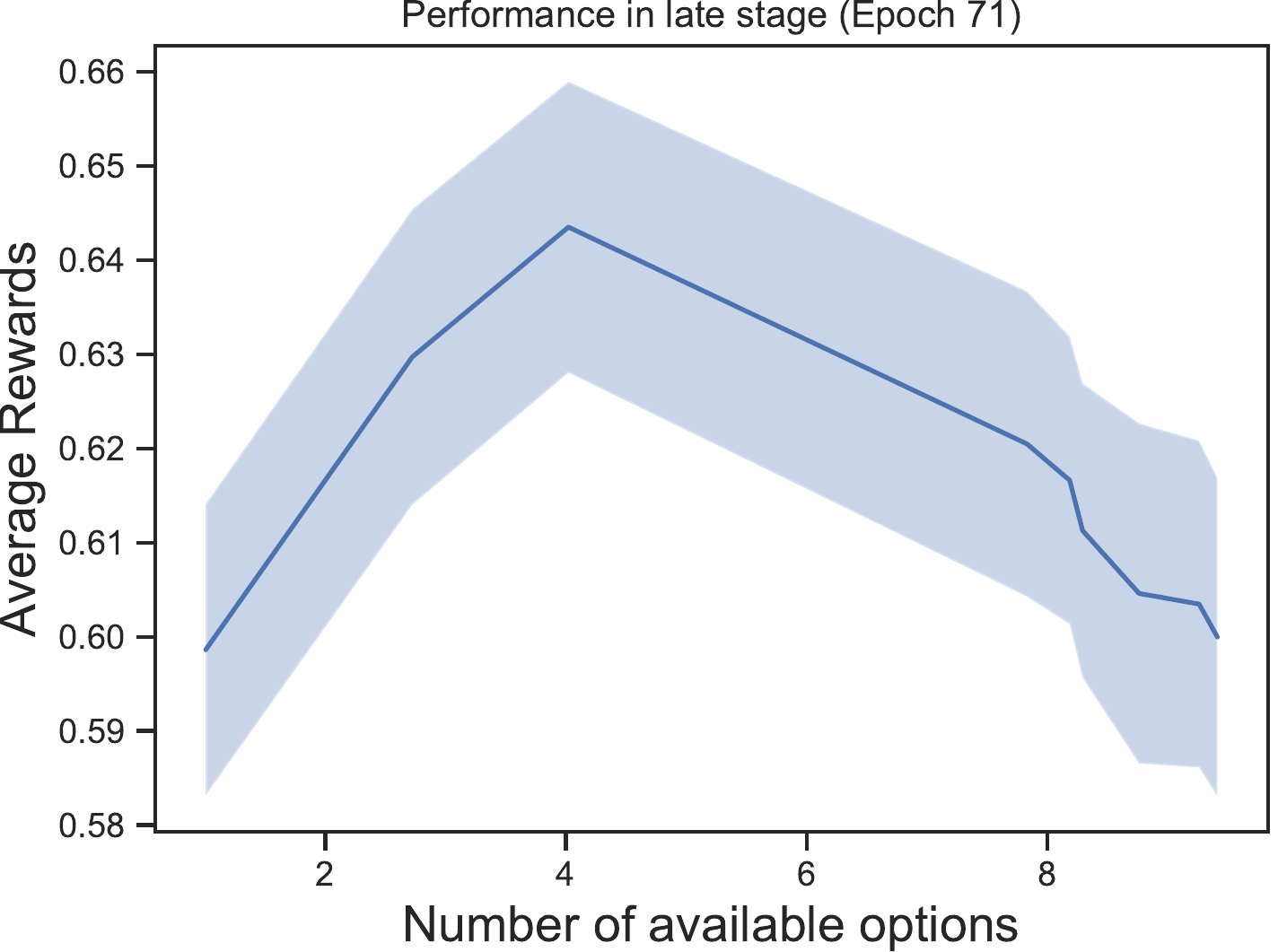}}
    \caption{\textbf{Interest functions and the branching factor}. During the initial stages of the learning process, allowing fewer options helps improve learning speed, whereas in the later stages, good solutions can still be obtained with a reasonable number of choices at each decision point.}
    \label{fig:branchingfactor}
\end{figure*}

\section{Preliminaries}
\textbf{Markov Decision Processes (MDPs).}
A finite, discrete-time  MDP \cite{puterman1995markov,SuttonBoook} is a tuple $ \langle {\cal S}, {\cal A}, r, P, \gamma \rangle $, where ${\cal S}$ is the set of states, ${\cal A}$ is the  set of actions, $r: {\cal S} \times {\cal A}\rightarrow \mathbb{R}$ is the reward function, $P:{\cal S} \times {\cal A} \rightarrow Dist({\cal S})$ is the environment transition probability function, and $\gamma \in [0,1)$ is the discount factor. At each time step, the learning agent perceives a state $S_t \in {\cal S}$, takes an action $A_t \in {\cal A}$ drawn from a policy $\pi : {\cal S} \times {\cal A} \rightarrow [0,1]$, and with probability $P(S_{t+1}|S_t,A_t)$ enters next state $S_{t+1}$, receiving a numerical reward $R_{t+1}=r(S_t, A_t)$ from the environment. The value function of policy $\pi$ is defined as: $V_\pi(s) = E_\pi[\sum_{t=0}^{\infty} \gamma^{t} R_{t+1} | S_0 = s]$ and its action-value function is: $Q_\pi(s,a) = E_\pi[\sum_{t=0}^{\infty} \gamma^{t} R_{t+1} | S_0 = s, A_0 = a]$.

\textbf{The Options framework.}
A Markovian option~\cite{sutton1999between} $\omega \in \Omega$ is composed of an \textit{intra-option policy} $\pi_\omega$, a termination condition $\beta_\omega: {\cal S} \rightarrow [0,1]$, where $\beta_\omega(s)$ is the probability of terminating the option upon entering state $s$, and an initiation set $I_\omega \subseteq {\cal S}$. In the \textit{call-and-return} option execution model, when an agent is in state $s$, it first examines the options that are available, i.e., for which $s\in I_\omega$. Let $\Omega(s)$ denote this set of available options. The agent then chooses $\omega\in \Omega(s)$ according to the policy over options $\pi_\Omega(s)$, follows the internal policy of $\omega$, $\pi_\omega$, until it terminates according to $\beta_\omega$, at which point this process is repeated. Note that $\Omega$ is the union of all sets $\Omega(s), \forall s$. The option-value function of $\omega \in \Omega(s)$ is defined as:
\begin{align*}
    Q_\Omega(s,\omega) = \sum_a \pi_{\omega}(a |  s) Q_U(s,\omega,a)\enspace ,
\end{align*} 
where  $Q_U: {\cal S} \times \Omega \times {\cal A} \rightarrow  \mathbb{R}$ is the value of executing primitive action $a$ in the context of state-option pair $(s,\omega)$:\begin{equation*}
    \begin{split}
        Q_U(s,\omega,a) = r(s,a) + \gamma \sum_{s'} P(s' | s,a) \\[-2ex]
         \times \left( (1 - \beta_{\omega}(s')) Q_\Omega(s',\omega) + \beta_{\omega}(s') \max_{\omega' \in \Omega(s')} Q_{\Omega}(s',\omega')\right)
    \end{split}
\end{equation*}
Note that if an option cannot initiate in a state $s$, its value is considered undefined.

\section{Interest-Option-Critic}
In~\citeauthor{sutton1999between} \shortcite{sutton1999between}, the focus is on discrete environments, and the notion of initiation set provides a direct analog of preconditions from classical planning. In large problems, options would be applicable in parts of the state space described by certain features. For example, an option of the form \textit{stop if the traffic light is red} would only need to be considered in states where a traffic light is detected.
Let $\mathbf{1}_{\omega}:{\cal S} \rightarrow \{0,1\}$ be the indicator function corresponding to set $I_\omega$: $\mathbf{1}_{\omega}(s)=1$ iff $s\in I_\omega$ and $0$ otherwise.

An \textit{interest function} $I_\omega : {\cal S} \times \Omega \rightarrow {\mathbb R}^{+}$ generalizes the set indicator function, with $I_\omega(s)>0$ iff $\omega$ can be initiated in $s$. A bigger value of $I_\omega(s)$ means the interest in executing $\omega$ in $s$ is larger. Note that, depending on how $I_\omega$ is parameterized, one could view the interest as a prior on the likelihood of executing $\omega$ in $s$. However, we will not use this perspective here, because our goal is to learn $I_\omega$. So, instead, we will choose a parameterized form for $I_\omega$ which is differentiable, in order to leverage the power of gradients in the learning process.

The value of $I_\omega$ modulates the probability of option $\omega$ being sampled in state $s$ by a policy over options $\pi_\Omega$, resulting in an \textit{interest policy over option} defined as:
\begin{equation}
    \pi_{I_{\omega}}(\omega | s) \propto I_{\omega}(s) \pi_{\Omega}(\omega | s)
\end{equation}
Note that this specializes immediately to usual initiation sets (where the interest is the indicator function). 

We will now describe an approach to learning options which includes interest functions. 
We propose a policy gradient algorithm, in the style of option-critic~\shortcite{bacon2017option}, based on the following result:   
\begin{theorem}
Given a set of Markov options with differentiable interest functions $I_{\omega,z}$, where $z$ is the parameter vector, the gradient of the expected discounted return with respect to $z$ at $(s, \omega)$ is:
\begin{align*}
  \sum_{s',\omega'} \hat{\mu}_{\Omega} (s', \omega' | s, \omega)  \beta_{\omega}(s') \frac{\partial \pi_{I_{\omega, z}}(\omega' | s')}{\partial z} Q_{\Omega}(s', \omega')
\end{align*}
where $\hat{\mu}_{\Omega} (s', \omega' | s, \omega)$ is the discounted weighting of the state-option pairs along trajectories starting from $(s, \omega)$ sampled from the distribution determined by $\pi_{I_{\omega,z}}$, $\beta_\omega$ is the termination function and $Q_\Omega$ is the value function over options corresponding to $\pi_{I_{\omega, z}}$.
\end{theorem}

The proof is in Appendix A.2.1 available on the project page\footnote{\url{https://sites.google.com/view/optionsofinterest}\label{site}}. We can derive policy gradients for intra-option policies and termination functions as in option-critic~\cite{bacon2017option} (see Appendix A.2.2, A.2.3) with the difference that the discounted weighting of state-option pairs is now according to the new option sampling distribution determined by $\pi_{I_{\omega,z}}(s)$. This is natural, as the introduction of the interest function should only impact the choice of option in each state. Pseudo-code of the interest-option-critic (IOC) algorithm  using intra-option Q-learning is shown in Algorithm~\ref{alg:ioc}. 

Intuitively, the gradient update to $z$ can be interpreted as increasing the interest in an option which terminates in states with good value. It links initiation and termination, which is natural. It is to be noted that the proposed gradient works at the level of the augmented chain; and not at the SMDP level. Implementing policy gradient at the SMDP level for the policy over options would entail performing gradient updates only upon termination, whereas using the augmented chain allows for updates throughout. Note that this approach does not appear in previous work to the best of our knowledge.

\section{Illustration}
In order to elucidate the way in which interest functions can help regulate the complexity of learning how to make decisions, we provide a small illustration of the idea.  Consider a point mass agent in a continuous 2-D maze, which starts in a  uniformly random position and must reach a fixed goal state. Consider a scalar threshold $ k \in [0,1]$, so that at any choice point, only options whose interest is at least $k$ can be initiated by the interest policy over options $\pi_{I_{\omega}}(\omega | s)$. The agent uses 16 options in total. Intuitively, an agent which has fewer option choices at a decision point should learn faster, since it has fewer alternatives to explore, but in the long run, this limits the space of usable policies for the agent. Fig.~\ref{fig:branchingfactor} confirms this trade-off between speed of learning and ultimate quality. Note that this trade-off holds the same way in planning as well (as discussed extensively in classical planning works).

\begin{algorithm}
		\caption{IOC with tabular intra-option Q-learning}
		\label{alg:ioc}
		\begin{algorithmic}
			\STATE Initialize policy over options $\pi_\Omega$
			\STATE Initialize $I_{\omega,z}$ parameterized by $z$ such that all options are available everywhere to some extent
			\STATE
			Initialize $\pi_{I_{\omega, z}}(\omega | s)$ as in Eq.(1)
			
			\STATE Set $s \longleftarrow s_0$ and $\omega$ at $s$ according to $\pi_{I_{\omega, z}}$
			\REPEAT
			\STATE Choose $a$ according to $\pi_{\omega,\theta}(a|s)$
			\STATE Take action $a$ in $s$, observe $s', r$

			\STATE Sample termination from $\beta{_\omega,\nu}(s')$
			\IF{$\omega$ terminates in $s'$}
			\STATE Sample $\omega'$ according to $\pi_{I_{\omega, z}}(\cdot|s')$
			\ELSE
            	\STATE $\omega'$ = $\omega$
            \ENDIF
            
			\textbf{1. Evaluation step:}
			\STATE $\delta \gets \, r - Q_U(s,\omega,a)$

			\STATE $\delta \gets \, r + \gamma(1-\beta_{\omega,\nu}(s')) Q_\Omega(s',\omega) + \gamma \beta_{\omega,\nu}(s') \max_{\omega'} Q_\Omega(s',\omega')$
			\STATE $Q_U(s,\omega,a) \gets \, Q_U(s,\omega,a) + \alpha \delta$
		
			\textbf{2. Improvement step}
			\STATE $\theta \gets \theta +\alpha_\theta \frac{\partial \log \pi_{\omega, \theta}(a| s)}{\partial \theta} Q_U(s, \omega,a)$
		    \STATE $\nu \gets \nu - \alpha_\nu \frac{\partial \beta_{\omega, \nu}(s')}{\partial \nu} (Q_\Omega(s', \omega) - V_\Omega(s')) \mbox{ where} V_{\Omega}(s') = \sum_\omega' \pi_{I_{\omega,z}}(\omega'|s') Q_\Omega(s',\omega')$
            \STATE $z \gets z + \alpha_z  \beta_{\omega,\nu}(s') \frac{\partial \pi_{I_{\omega, z}}(\omega' | s')}{\partial z} Q_{\Omega}(s', \omega')$
            \newline
            $s\leftarrow s'$
			\UNTIL{$s'$ is a terminal state}
		\end{algorithmic}
\end{algorithm}

\section{Experimental Results}
\label{sec:experiments}
We now study the empirical behavior of IOC in order to answer the following questions: (1) are options with interest functions useful in a single task; (2) do interest functions facilitate learning reusable options and, (3) do interest functions provide better interpretability of the skills of an agent. A link to the source code for all experiments is provided on the project page.
\begin{figure*}[h!]
    \centering
    \subfigure[]{\label{fig:FR_UniformPiO}\includegraphics[width=0.31\textwidth]{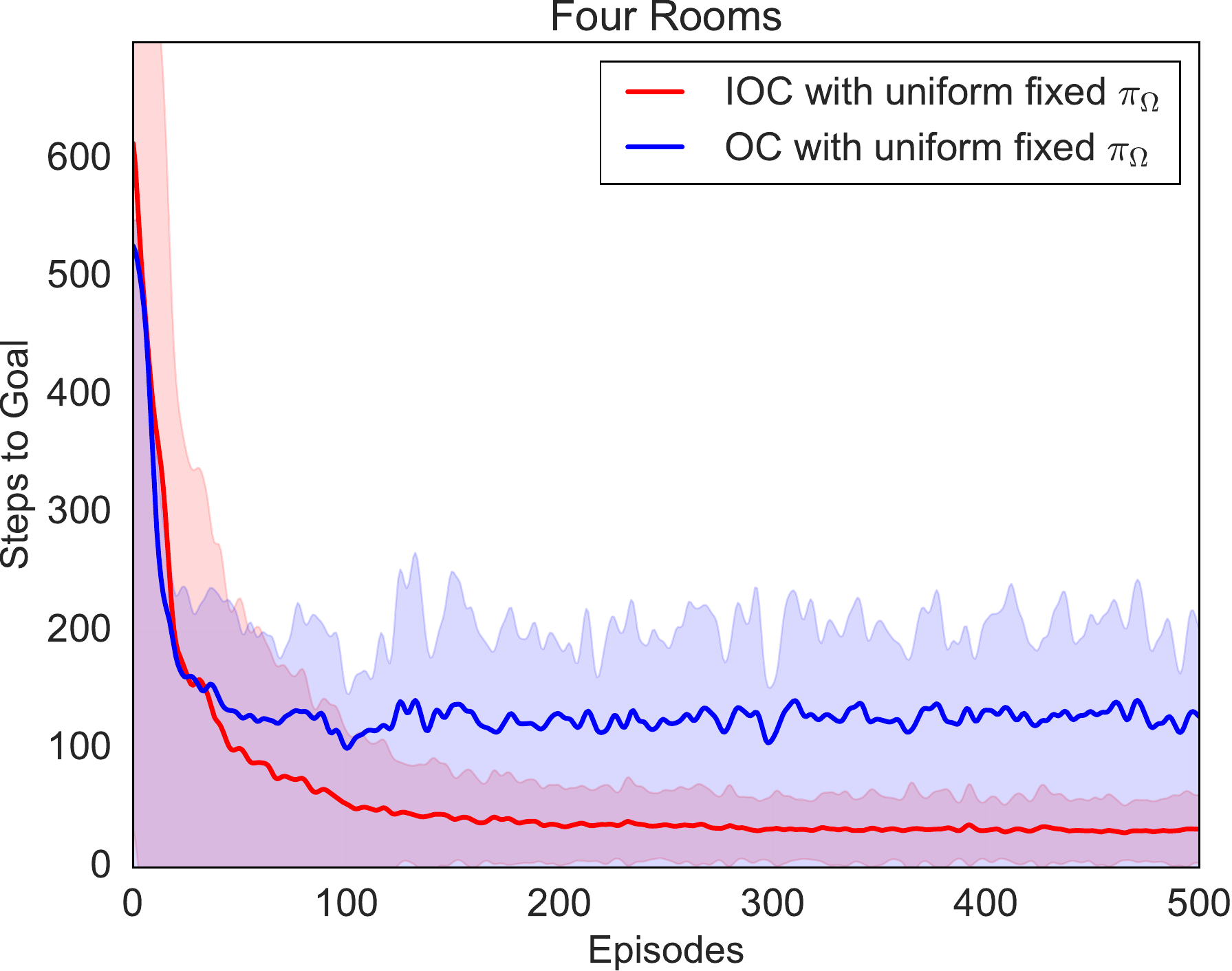}}
    \subfigure[]{\label{fig:TMaze_UniformPiO}\includegraphics[width=0.32\textwidth]{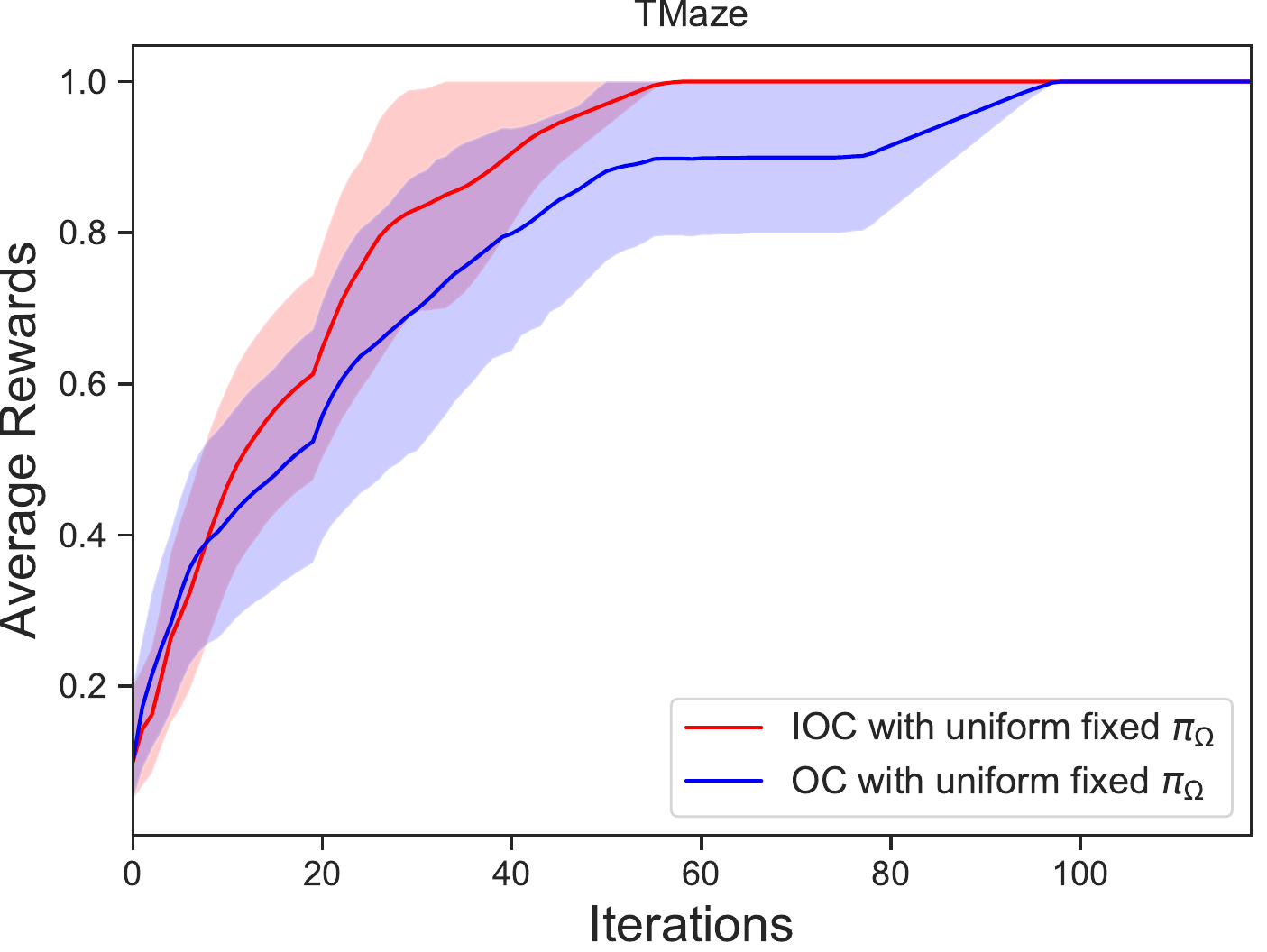}}
    \subfigure[]{\label{fig:Miniworld_UniformPiO}\includegraphics[width=0.32\textwidth]{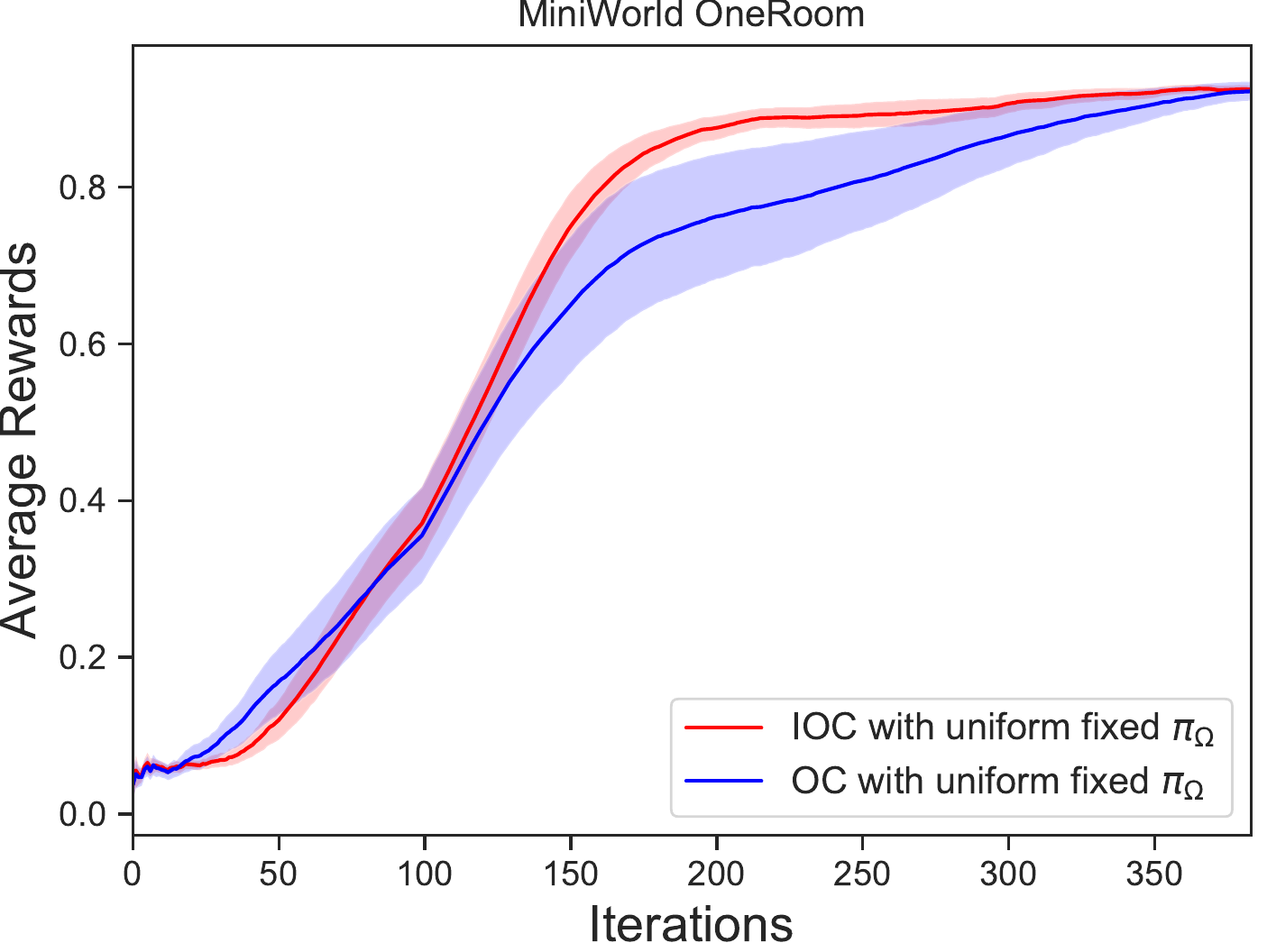}}
    \label{fig:MisspecifiedPiO}
    \caption{\textbf{Learning options with interest reshapes a uniform fixed policy over options} in a single task,  in 3 different domains: (a) tabular Four rooms, (b) continuous control in TMaze, and (c) 3D visual Miniworld task. IOC  outperforms OC, indicating the utility of interest functions.}
\end{figure*}

\subsection{Learning in a single task}
\label{sec:Q1}
To analyze the utility of interest functions when learning in a single task, consider a given, fixed policy over options, either specified by a just-in-time planner or via human input. This setup allows us to understand the impact of interest functions alone in the learning process.

\begin{figure*}[h]
    \centering
    \subfigure[$I_{\omega_1}$]{\label{fig:Final_IntFn_opt0}\includegraphics[width=0.22\textwidth]{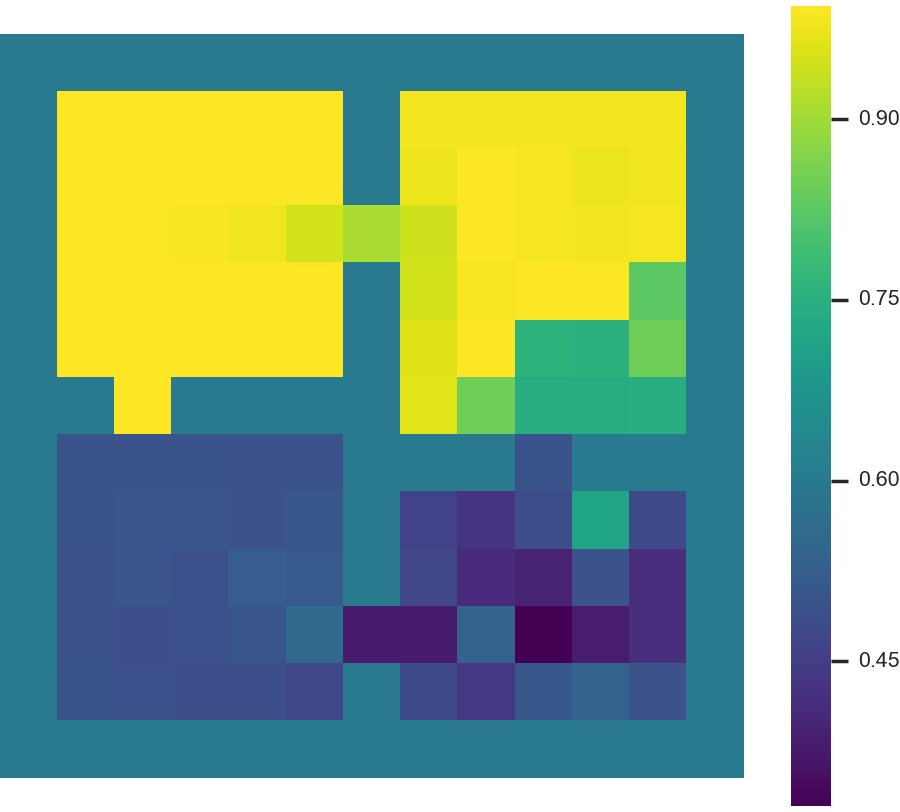}}
    \subfigure[$I_{\omega_2}$]{\label{fig:Final_IntFn_opt1}\includegraphics[width=0.22\textwidth]{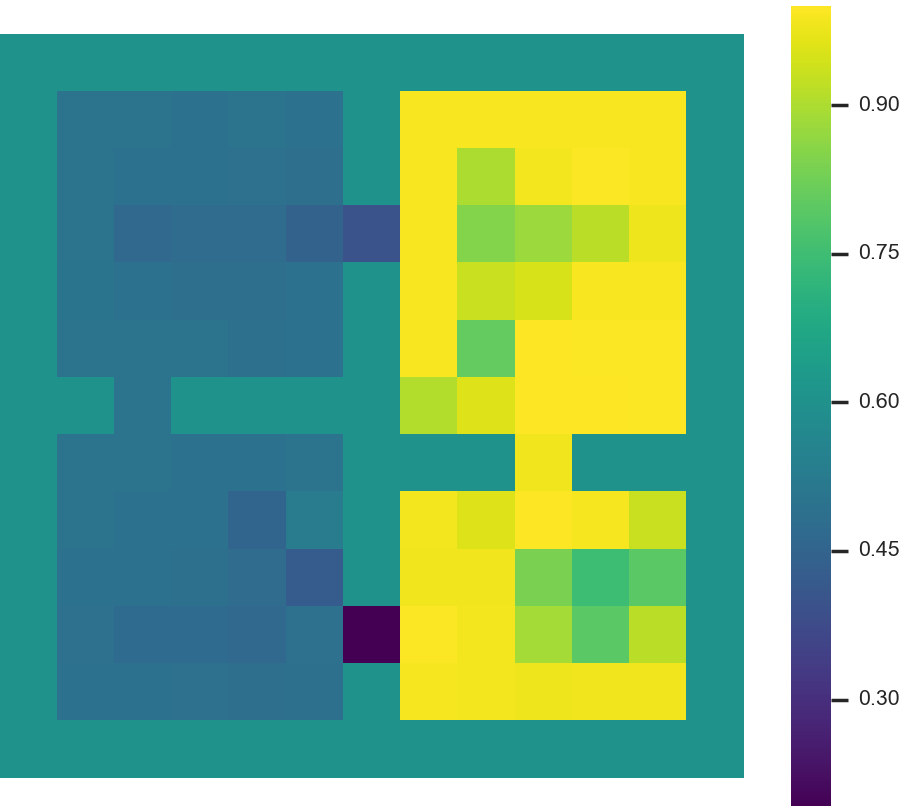}}
    \subfigure[$I_{\omega_3}$]{\label{fig:Final_IntFn_opt2}\includegraphics[width=0.22\textwidth]{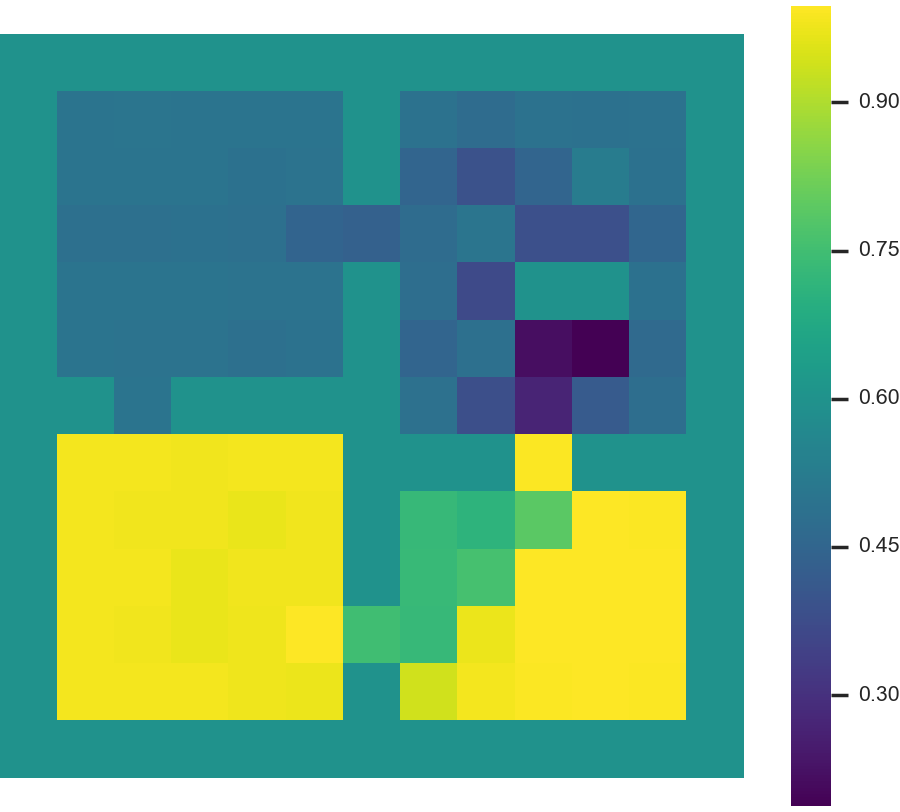}}
    \subfigure[$I_{\omega_4}$]{\label{fig:Final_IntFn_opt3}\includegraphics[width=0.22\textwidth]{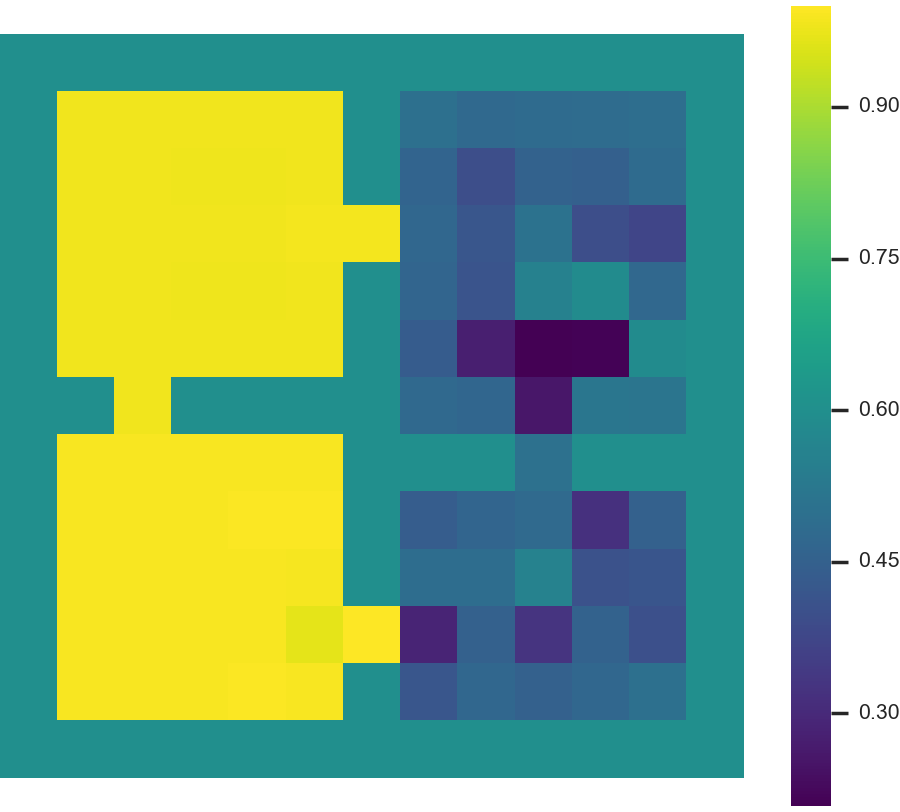}}
    \caption{\textbf{Visualization of Interest Functions ($I_{\omega}$)} at the end of $500$ episodes with the goal in the east hallway. Brighter colors represent higher values. Options learned with interest functions emerge with specific interest in different regions of the state.}
    \label{fig:FRIOCinterestfunction}
\end{figure*}

\textbf{\emph{Four rooms (FR)}} We first consider the classic FR domain~\shortcite{sutton1999between} (Fig.~\ref{fig:Final_IntFn_opt0}). The agent starts at a uniformly random state and there is a \textit{goal} state in the bottom right hallway ($\gamma=0.9$). With probability $1/3$, the agent transitions randomly to one of the empty adjacent cells instead of the desired movement. The reward is $+50$ at the goal and $0$ otherwise. We used $4$ options, whose intra-option policies were parameterized with Boltzmann distributions, and termination and interest functions represented as linear-sigmoid functions. Options were learned using either IOC or OC with tabular intra-option Q-learning, as described in Algorithm~\ref{alg:ioc}. Learning proceeds for $500$ episodes, with a maximum of $2000$ time steps allowed per episode. Additional details are provided in Appendix A.3.1.

\textbf{Results:} Fig.~\ref{fig:FR_UniformPiO} shows the steps to goal for both OC and IOC, averaged over $70$ independent runs. The IOC agent performs better than OC agent by roughly $100$ steps. One potential reason for the improvement in IOC is that options become specialized to different regions of the state-space, as can be seen in Fig.~\ref{fig:FRIOCinterestfunction}. We also observe that the termination functions (which were initialized to $0$) naturally become coherent with the interest functions learned, and are mostly room specific for each option (see appendix Fig. A1). On the other hand, options learned by OC do not show such specialization and terminate everywhere (see appendix Fig. A1). These results demonstrate that the IOC agent is not only able to correct for the given higher level policy, but also, leads to more understandable options as a side effect. 

\begin{figure*}[h]
    \centering
    \subfigure[]{\label{fig:TMazeTransferPerf}\includegraphics[width=0.32\textwidth]{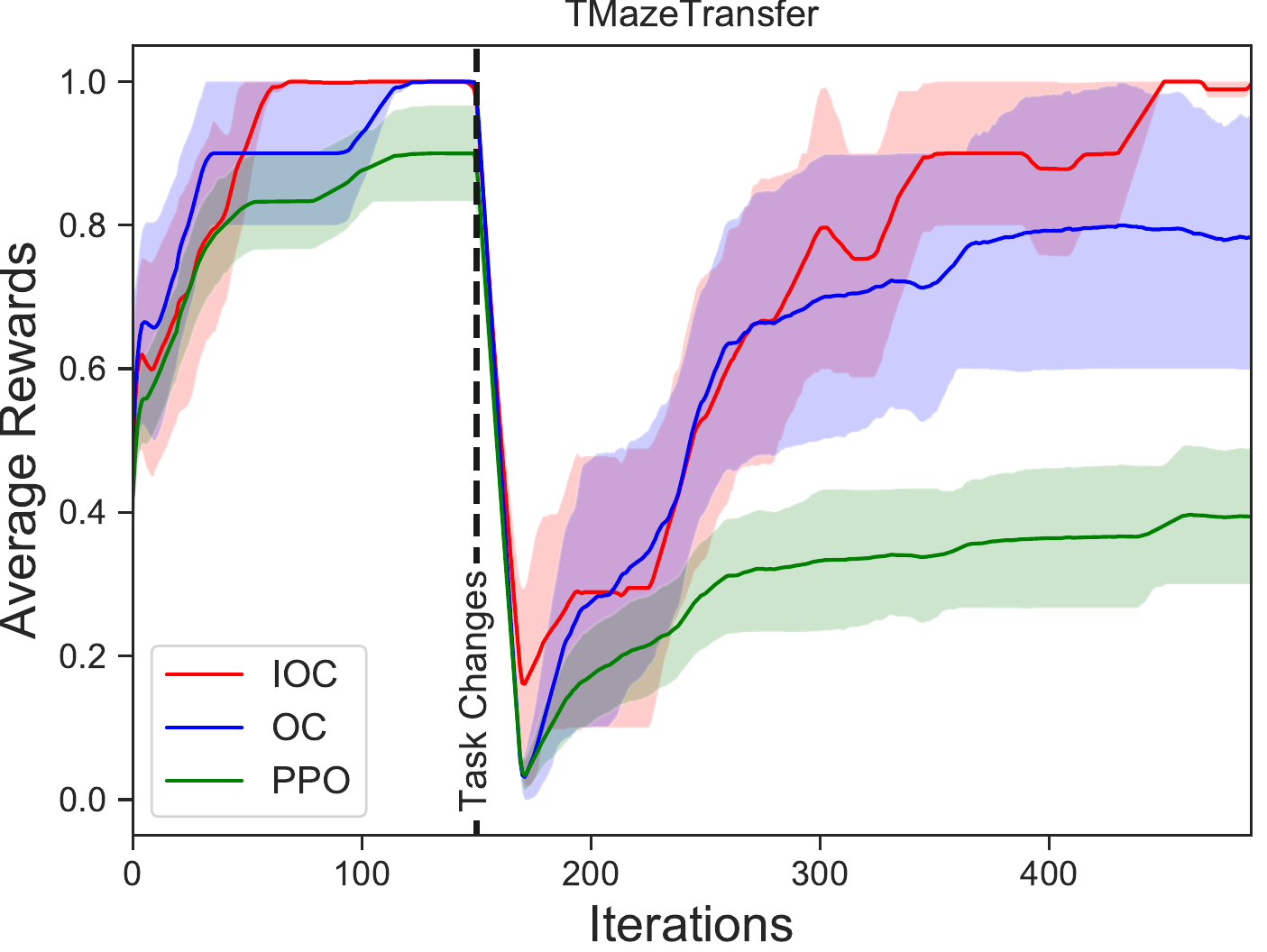}}
    \subfigure[]{\label{fig:MiniworldTransfer}\includegraphics[width=0.32\textwidth]{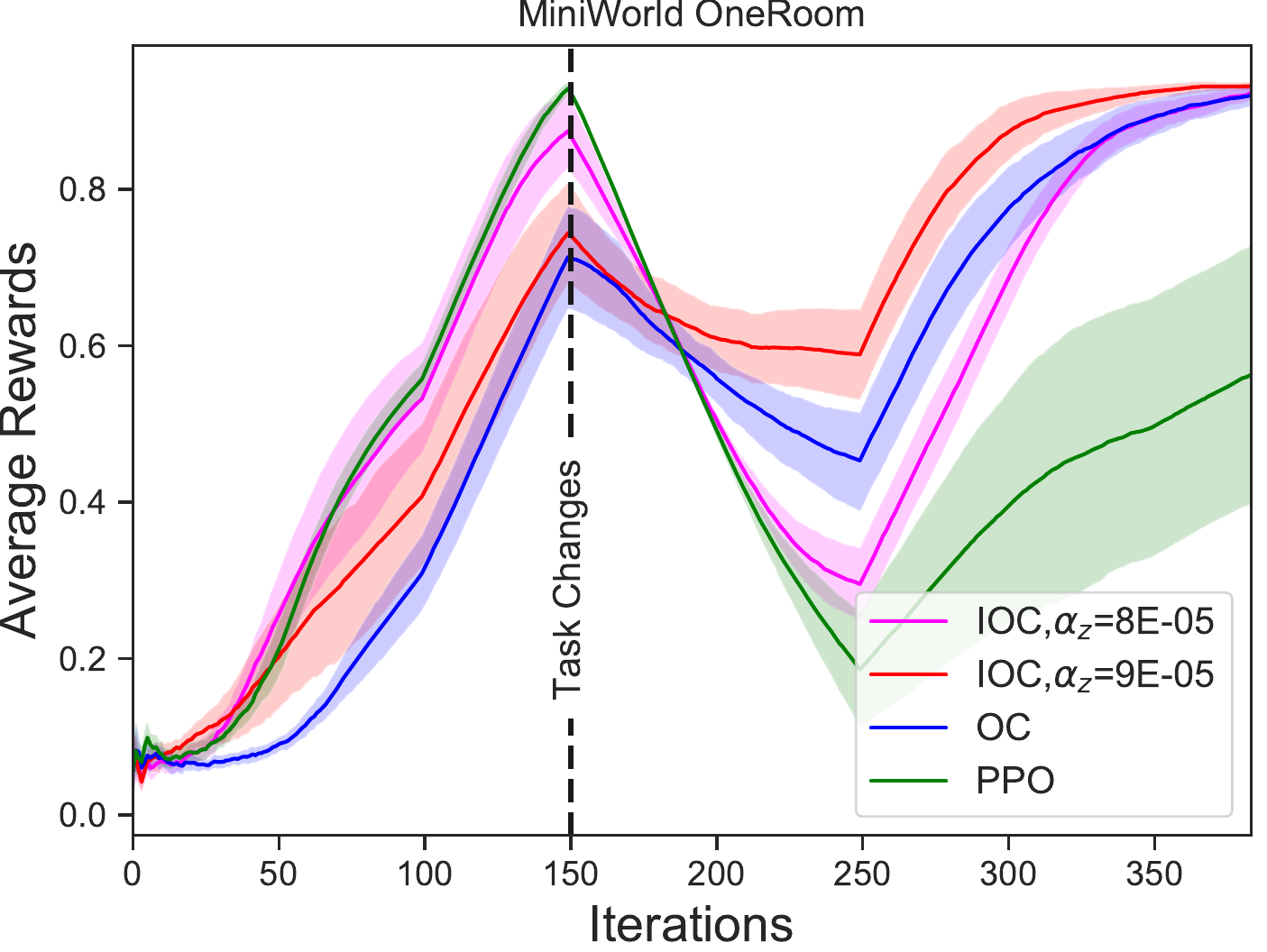}}
    \subfigure[]{\label{fig:HalfCheetahTransfer}\includegraphics[width=0.32\textwidth]{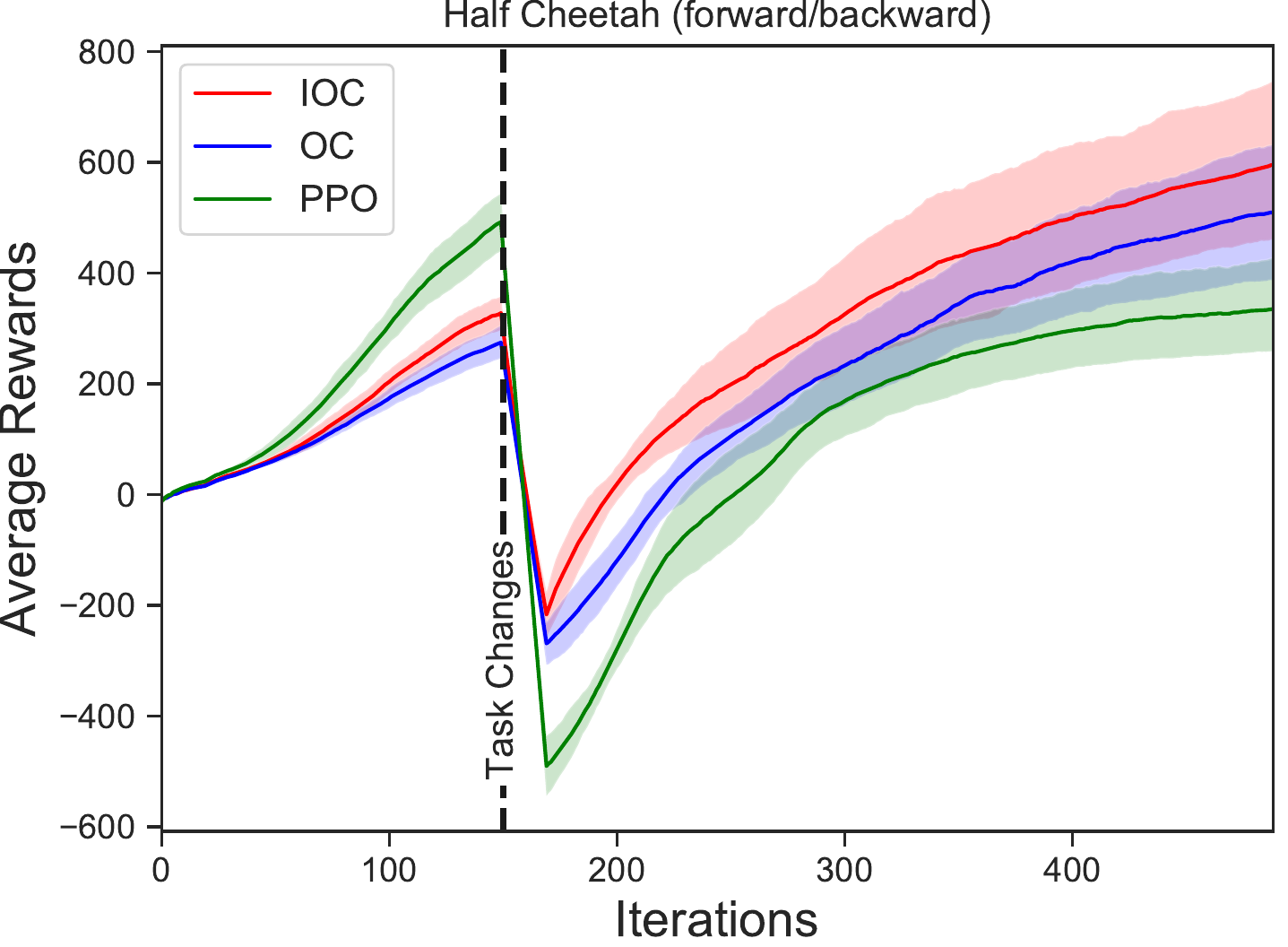}}
    \quad
    \subfigure[\small ]{\label{fig:TMazeTwoGoals}\includegraphics[width=0.155\textwidth]{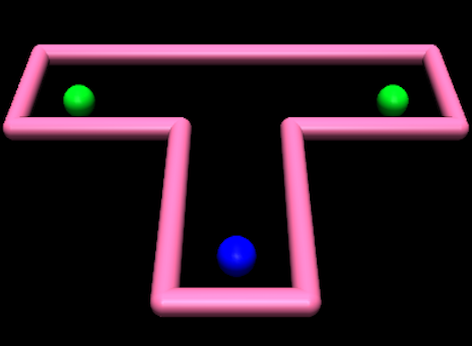}}
    \subfigure[\small ]{\label{fig:TMazeSingleGoal}\includegraphics[width=0.155\textwidth]{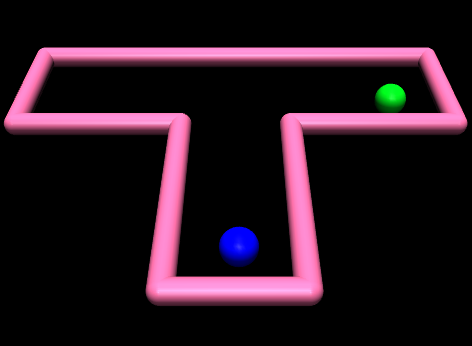}}
    \subfigure[\small ]{\label{fig:Navigate_RedBlock}\includegraphics[width=0.15\textwidth]{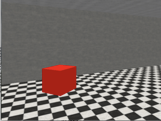}}
    \subfigure[\small ]{\label{fig:Naivate_BlueBlock}\includegraphics[width=0.15\textwidth]{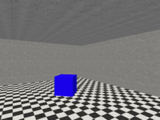}}
    \subfigure[\small ]{\label{fig:HCTask1}\includegraphics[width=0.15\textwidth]{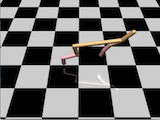}}
    \subfigure[\small ]{\label{fig:HCTask2}\includegraphics[width=0.15\textwidth]{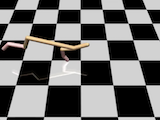}}
    \caption{\textbf{Transfer in continual learning setting:} \textit{Continuous Control in TMaze:} (d) The point mass agent ({blue}) has two goals ({green}) both resulting in a reward of $+1$. After $150$ iterations, the goal that is most visited is removed (e). (a) illustrates that IOC converges fastest in the first few iterations. After the task change, IOC suffers the least in terms of immediate loss in performance and gets the best final score. \textit{Visual navigation in Miniworld:} requires  the agent to go to a randomly placed {red} box in a closed room. After $150$ episodes the color of the box changes and the agent has to navigate to a unseen {blue} box (g). IOC quickly adapts to the change (b) indicating that harnessing options learned from the old task speeds up learning in the new task. \textit{Locomotion in HalfCheetah:} The cheetah is rewarded for moving forward as fast as possible during the first 150 iterations, after which it is rewarded for going backwards as fast as possible.}
    \label{fig:transferlearning}
\end{figure*}

\textbf{\emph{TMaze}} Next, we illustrate the learning and use of interest functions in the non-linear function approximation setting, using simple continuous control tasks implemented in Mujoco \cite{todorov2012mujoco}. A point mass agent (blue) is located at the bottom end of a T-shaped maze ($0$, $-0.1$) and must navigate to within $0.1$ of the goal position ($0.3$, $0.3$) at the right end of the maze (green location) (Fig.~\ref{fig:TMazeSingleGoal}). The state space consists of the $x, y$ coordinates of the agent and the action space consists of force applied in the $x, y$ directions. We use a uniform fixed policy over options for both IOC and OC. We reuse the Proximal Policy Option-Critic (PPOC) algorithm \cite{ppoc_klissarov}\footnote{In this work we name this algorithm OC for option-critic} and add a 2-layer network with sigmoid outputs to compute the interest functions. However, we do correct the implementation of the gradient of the policy over options which has been overlooked in that work. The remaining update rules are consistent with Algorithm~\ref{alg:ioc}. Complete details about the implementation and hyper-parameter search are provided in Appendix A.3.2.

\begin{figure*}[ht!]
    \centering
    \subfigure{
    \hspace{-40pt}
    \begin{tikzpicture}
        \node[anchor=south west,inner sep=0] at (0,0) {\includegraphics[width=0.75\textwidth]{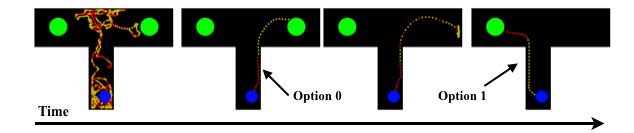}};
        \node[] (a) at (-0.5,1) {(a)};
    \end{tikzpicture}
    \label{fig:tmaze_trajectories}
    }\\[-1.8ex]
    \subfigure{
    \hspace{-40pt}
    \begin{tikzpicture}
        \node[anchor=south west,inner sep=0] at (0,0) {\includegraphics[width=0.75\textwidth]{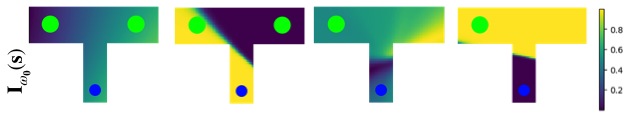}};
        \node[] (b) at (-0.5,1) {(b)};
    \end{tikzpicture}
    \label{fig:tmaze_interest_option0}
    }\\[-2.0ex]
    \subfigure{
    \hspace{-40pt}
    \begin{tikzpicture}
        \node[anchor=south west,inner sep=0] at (0,0) {\includegraphics[width=0.75\textwidth]{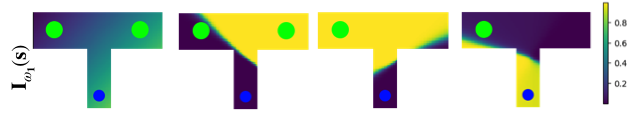}};
        \node[] (c) at (-0.5,1) {(c)};
    \end{tikzpicture}
    \label{fig:tmaze_interest_option1}
    }\\[-1.5ex]
    \subfigure{\label{fig:tmaze_sequence}\includegraphics[width=0.75\textwidth]{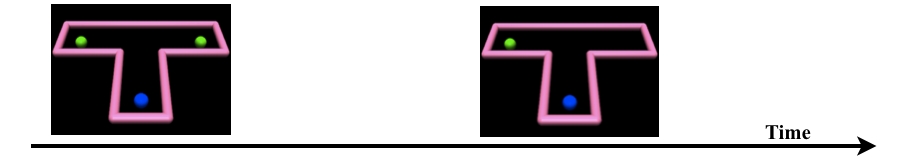}}
    \caption{\textbf{Qualitative analysis of IOC in TMaze} illustrates the options and their interest over time. \textbf{Row a} depicts sampled trajectories with option $0$ indicated by red dots and option $1$ by yellow dots. \textbf{Row b \& c} depict the interest of each option. At the end of task 1, the interest of option $0$ emerges in the lower diagonal (row b) whereas option $1$ is interested in a different region (row c). As the task changes, the interest functions adapt (col $3$).}
    \label{fig:qualitativeanalysisTMaze}
\end{figure*}
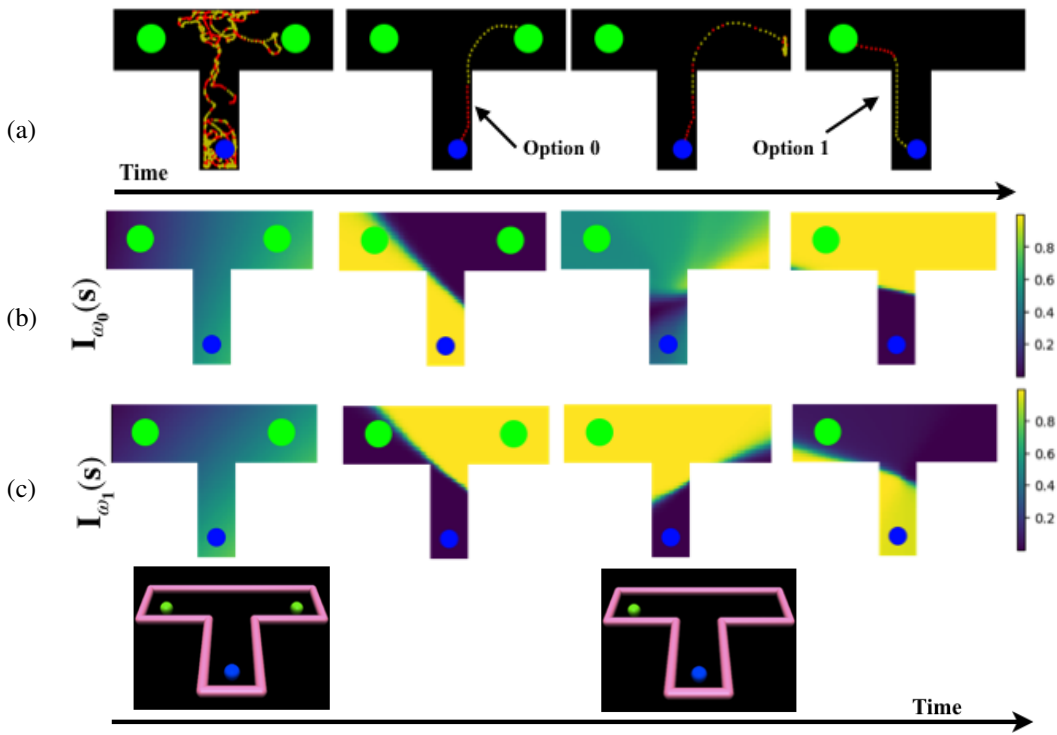

\textbf{Results:} We report the average performance over $10$ independent runs. The IOC agent  is able to converge in almost half the time steps needed by the OC agent. Potentially, interest functions in the IOC agent provide an attention mechanism and thus facilitates learning options which are more diverse (see Fig.~\ref{fig:qualitativeanalysisTMaze} for evidence). A deeper analysis of interest functions learned in this domain is deferred to subsequent sections.

\textbf{\emph{MiniWorld}} We also explore learning in more complex $3$D first-person visual environment from the MiniWorld framework~\cite{gym_miniworld}. We use the \textit{Oneroom} task where the agent has to navigate to a randomly placed red block in a closed room (Fig.~\ref{fig:Navigate_RedBlock}). This requires the agent to turn around and scan the room to find the red box, then navigate to it. 

The observation space is a $3$-channel RGB image and the action space consists of $8$ discrete actions. At the start of each episode, the red box is placed randomly. The episode terminates if the agent reaches the box or max of $180$ time steps is reached. We used the DQN architecture of~\cite{mnih2015human}. See Appendix A.3.3 for details about implementation and hyper-parameters.

\textbf{Results:} The IOC agent is able to converge much faster ($~100$ iterations) than the OC agent with a given uniform policy over option (Fig.~\ref{fig:Miniworld_UniformPiO}). The performance is averaged across $10$ runs.

Based on these experiments, IOC provides improvement in performance consistently across a range of tasks, from the simple four-rooms domain to complex visual navigation tasks such as MiniWorld, indicating the utility of learning interest functions. 

\subsection{Option reusability}
\label{sec:Q2}
One of the primary reasons for an agent to autonomously learn options is the ability to generalize its knowledge quickly in new tasks. We now evaluate our approach in settings where adaptation to changes in the task is vital.

\textbf{\emph{TMaze}} The point mass agent starts at the bottom of the maze, with two goal locations (Fig.~\ref{fig:TMazeTwoGoals}), both giving a reward of $+1$. After $150$ episodes, the goal that has been visited the most is removed and the agent has to adapt its policy to the remaining goal available (Fig.~\ref{fig:TMazeSingleGoal}). Both OC and IOC learn $2$ options. We use a softmax policy over options for both IOC and OC, which is also learned at the same time. 

\textbf{Results:} In the initial phase, the difference in performance between IOC and the other two agents (OC and PPO) is striking (Fig.~\ref{fig:TMazeTransferPerf}): IOC converges twice as fast. Moreover, when the most visited goal is removed and adaptation to the task change is required, the IOC agent is able to explore faster and its performance declines less. This suggests that the  structure learned by IOC provides more generality. At the end of task 2, IOC recovers its original performance, whereas PPO fails to recover during thee allotted learning trials.

\textbf{\emph{MiniWorld}} Initially, the agent is tasked to search and navigate to a randomly placed red box in one closed room (Fig.~\ref{fig:Navigate_RedBlock}). After $150$ episodes, the agent has to adapt its skills to navigate to a randomly located blue box (Fig.~\ref{fig:Naivate_BlueBlock}) which it has never seen before. Here, the policy over options as well as all the option components are being learned at the same time.

\textbf{Results:} The IOC agent outperforms both OC and PPO agents when required to adapt to the new task (Fig.~\ref{fig:MiniworldTransfer}). This result indicates that the options learned with interest functions are more easily transferable. The IOC agent is able to adapt faster to unseen scenarios.

\textbf{\emph{HalfCheetah}} We also study adaptation in learning a complex locomotion task for a planar cheetah. The initial configuration of this environment follows the standard HalfCheetah-v1 from OpenAI's Gym: the agent is rewarded for moving forward as fast as possible. After $150$ iterations, we modify the reward function so that the agent is now encouraged to move backward as fast as possible \cite{finn2017model}. 

\textbf{Results:} PPO outperforms both OC and IOC in the initial task. However, as soon as the task changes, IOC reacts in the most efficient way and converges to the highest score after $500$ iterations (Fig.~\ref{fig:HalfCheetahTransfer}). As seen consistently in all the environments, IOC generalizes much better over tasks, whereas PPO seems to overfit to the first task and generalizes poorly when the task changes.

In all our experiments, we notice that interest functions result in option specialization, which leads to both reusability and \textit{adaptability} (i.e. an option may get slightly tweaked), especially in the complex tasks.

\subsection{Option interpretability}
\label{sec:interpretability}
To gain a better understanding of the agent's behavior, we visualize different aspects of the learning process in several tasks.

\textbf{\emph{TMaze}}  We visualize the interest functions learned in TMaze (Fig.~\ref{fig:qualitativeanalysisTMaze}). Initially, the interest functions are randomized. At the end of the first task, the interest function for option $0$ specializes in the lower diagonal of the state-space (Fig.~\ref{fig:tmaze_interest_option0}), whereas option $1$'s interest function is completely different (Fig.~\ref{fig:tmaze_interest_option1}). When the task changes, the options readjust their interest. Eventually, the interest functions for the two options automatically specialize in \textit{different} regions of the state space (last column of Fig.~\ref{fig:tmaze_interest_option1} \& ~\ref{fig:tmaze_interest_option0}). Fig.~\ref{fig:tmaze_trajectories} illustrates the agent trajectories at different time instances, where the yellow and red dots indicate the two different options during the trajectory. A visualization of the emergence of interest functions during the learning process is also available on the project page (see ~\ref{site}). In contrast, the options learned by the OC agent are employed everywhere and have not specialized as much (see Appendix Fig. A2).

\begin{figure*}[ht]
    \centering
    \includegraphics[width=0.7\textwidth]{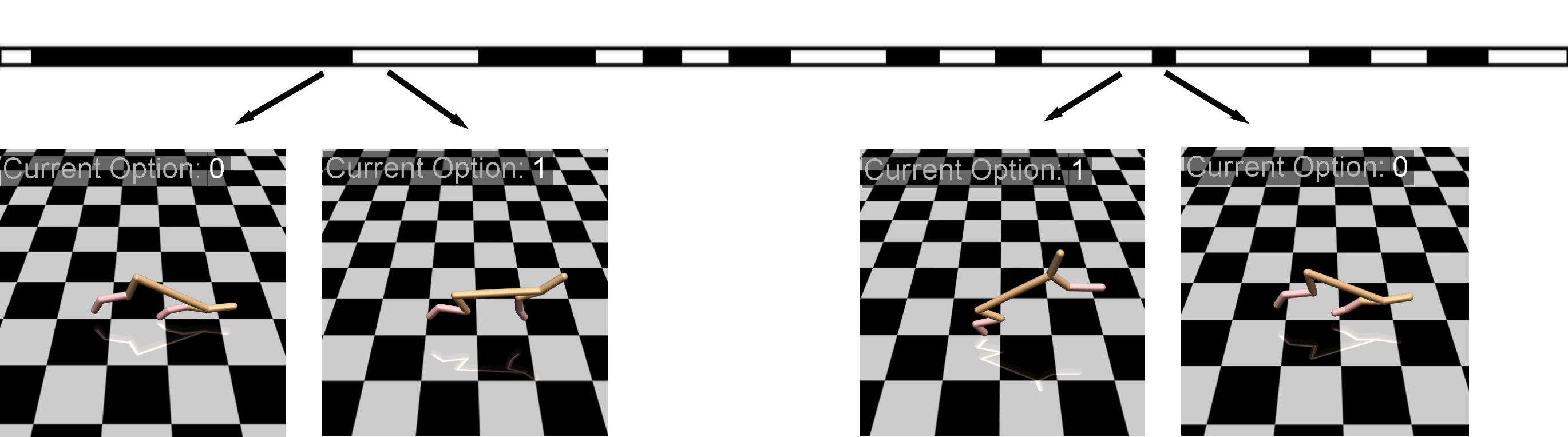}
    \caption{\textbf{Timeline of options used by IOC agent in HalfCheetah} where each option is represented by a distinct color (black \& white). Two distinct options are learned during the task of moving forward; option $0$  moves forward by dragging the limbs whereas option $1$ takes larger hopped steps; see accompanying videos.}
    \label{fig:halfcheetahskills}
\end{figure*}

\textbf{\emph{HalfCheetah}} We analyze the skills learned in HalfCheetah. During the task of moving forward as fast as possible, the IOC agent employs option $0$ to move forward by dragging its limbs, and option $1$ to take much larger hopped steps (Fig.~\ref{fig:halfcheetahskills}).  Fig.~\ref{fig:halfcheetahskills} demonstrates the emergence of these very distinct skills and the agent's switching between them across time. Additionally, we analyzed each option at the end of task $2$ in which the agent was rewarded for moving backward. Option $0$ now specializes in moving forward while option $1$ focuses on moving backward. This is nice, as the agent preserves some ability to now solve both tasks. OC doesn't learn options which are as distinct, and both options end up going backward and overfitting to the new task (see accompanying videos (~\ref{site})). 

\begin{figure*}[ht]
    \centering
    \includegraphics[width=0.8\textwidth]{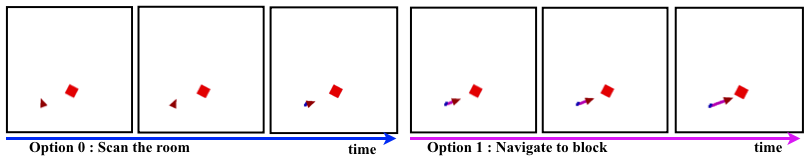}
    \caption{\textbf{3D Navigation in MiniWorld} with the top view of the environment and the agent as the red arrow. We show sequence of frames from a sampled trajectory. Option 0 scans the environment to locate the block. Option 1 learns to directly navigate to the block, once located. Please see accompanying videos.}
    \label{fig:miniworldskills}
\end{figure*}

\textbf{\emph{MiniWorld}} We visualize the skills acquired by inspecting the agent's behavior at the end of first task. The IOC agent has learned two distinct options: option $0$ scans the surroundings, whereas option $1$ is used to directly navigate towards the block upon successfully locating it (Fig.~\ref{fig:miniworldskills}). During task 2, option $1$ is being harnessed primarily to move forward, whereas option $0$ is employed when jittery motion is involved, such as turning and scanning. 

\section{Related Work}
Temporal abstraction in RL has a rich history \cite{parr1998reinforcement,thrun1995finding,dayan1993feudal,dietterich2000hierarchical,mcgovern2001automatic,menache2002q,stolle2002learning}. Options in particular have been shown to speed up convergence both empirically \cite{precup2000temporal} and theoretically \cite{mann2014scaling}. Constructing such temporal abstractions automatically from data has also been tackled extensively, and with some success~\cite{konidaris2011autonomous,mann2015approximate,kulkarni2016hierarchical,mankowitz2016adaptive,bacon2017option,machado2017laplacian}.  While some of the approaches require prior knowledge, have a fixed time horizon for partial policies \cite{vezhnevets2017feudal}, or use intrinsic rewards \cite{kulkarni2016hierarchical}, \citeauthor{bacon2017option}~\shortcite{bacon2017option} provides an end-to-end differentiable approach without needing any sub-goals or intrinsic motivation. 
We generalize their policy gradient approach to learn interest functions.
While we use rewards in our gradient-based algorithm, our qualitative analysis also indicates some clustering of states in which a given option starts, as in~\cite{lakshminarayanan2016option,bacon2013bottleneck,mannor2004dynamic}. Our approach is closely related in motivation to \citeauthor{mankowitz2016adaptive} \shortcite{mankowitz2016adaptive}. However, our method does not make assumptions about a particular structure for the initiation and termination functions (except smoothness). 

Initiation sets were an integral part of~\citeauthor{sutton1999between}~\shortcite{sutton1999between} and provide a way to control the complexity of the process of exploration and planning with options. This aspect of options has been ignored since, including in recent works~\cite{bacon2017option,harb2017waiting,harutyunyan2019termination,harutyunyan2019per} because there was no elegant way to learn initiation sets. We address this \textit{open problem} by generalizing initiation sets to differentiable interest functions. Since an interest function is a component of an option, it can be transferred once learned.
\section{Discussion and Future Directions}
We introduced the notion of interest functions for options, which generalize initiation sets, with the purpose of controlling search complexity. We presented a policy gradient-based method to learn options with interest functions, using general function approximation. Because the learned options are specialized, they are able to both learn faster in a single task and adapt to changes much more efficiently than options which initiate everywhere. Our qualitative results suggest that the interest function could be interpreted as an \textit{attention mechanism}  (see Appendix Fig. A4). To some extent, the interest functions learnt are able to override termination degeneracy as well (only one option being active all the time, or options switching often) although our approach was not meant to tackle that problem directly. Exploring further the interaction of initiation and termination functions, and imposing more coordination between the two, is an interesting topic for future work.

In our current experiments, the agent optimizes a single external reward function. However, the same algorithm could be used with intrinsic rewards as well. 

We did not explore in this paper the impact of interest functions in the context of planning.  However, given the intuitions from classical planning, learning models for options with interest functions could lead to better and faster planning, which should be explored in the future.

Finally, other ways of incorporating interest functions into the policy over options would be worth considering, in order to consider only choices over few options at a time.

\section{Acknowledgments}
The authors would like to thank NSERC \& CIFAR for funding this research; Emmanuel Bengio, Kushal Arora for useful discussions throughout this project; Michael Littman, Zafarali Ahmed, Nishant Anand, and the anonymous reviewers for providing critical and constructive feedback.

\fontsize{9.0pt}{10.0pt} \selectfont
\bibliographystyle{aaai}
\bibliography{main}
\end{document}